\documentclass[letterpaper]{article} 
\usepackage[preprint]{aaai2027}  
\usepackage[hyphens]{url}  
\usepackage{graphicx} 
\def\UrlFont{\rm}  
\usepackage{natbib}  
\usepackage{caption} 
\usepackage{amsmath}
\usepackage{amssymb}
\usepackage{algorithm}
\usepackage{algorithmic}

\usepackage{newfloat}
\usepackage{listings}
\DeclareCaptionStyle{ruled}{labelfont=normalfont,labelsep=colon,strut=off} 
\floatstyle{ruled}
\newfloat{listing}{tb}{lst}{}
\floatname{listing}{Listing}

\usepackage{booktabs}
\usepackage{multirow}

\title{CT-PrepAgent: Bounded Policy and Controlled Execution for Adaptive CT Data Preparation}

\author{
    Xiaolin Fan\textsuperscript{\rm 1},
    Yue Pei\textsuperscript{\rm 1},
    Yingying Zhang\textsuperscript{\rm 1},
    Haogang Zhu\textsuperscript{\rm 1}
}

\affiliations{
    \textsuperscript{\rm 1}
    Beihang University, Beijing, China\\
    fanxiaolin@buaa.edu.cn, peiyue@buaa.edu.cn,
    zhangyingying@buaa.edu.cn, haogangzhu@buaa.edu.cn
}

\begin{document}

\maketitle

\begin{abstract}
Heterogeneous computed tomography (CT) acquisitions and diverse downstream task requirements limit the transferability of fixed data preparation workflows across data sources and tasks. Existing approaches typically rely on manually designed or dataset-specific rules, making it difficult to accommodate changes in acquisition conditions and analytical objectives without manual intervention. Large language model (LLM)-based agents have shown promise for automating medical workflows, yet their potential for adaptive CT data preparation remains largely unexplored. To bridge this gap, we propose CT-PrepAgent, which enables adaptive CT data preparation through a bounded policy and controlled deterministic execution. Deterministic inspection constructs structured data--task profiles, from which a policy decides an eligible DICOM series or predefined preprocessing profile, while the controlled execution flow guards, resolves, executes, and verifies the decision with bounded recovery when enabled and safe quarantine otherwise. Across three public CT segmentation tasks, CT-PrepAgent derived data-task adaptive preprocessing decisions and achieved the highest macro-average Dice. On two private raw-DICOM cohorts, CT-PrepAgent increased verified output yield from 61.7\% to 70.0\% and yielded similar registration metrics on common verified outputs. Controlled fault and replay tests validate bounded recovery, safe quarantine, and policy-free replay under tested fault and drift settings.

\end{abstract}


\section{Introduction}
Preparing computed tomography (CT) data for downstream task analysis requires more than format conversion \cite{jin2024preprocessing}. Decisions concerning series selection, spatial geometry, intensity transformation, and quality control determine which information reaches the downstream task and can substantially affect task performance \cite{murray2024lazy,shin2023intensity}. These decisions become difficult to standardize because CT data vary across acquisition protocols, reconstruction settings, anatomical regions, and storage organizations, while segmentation, registration, and other tasks impose different input requirements \cite{denner2023efficient,buddenkotte2024ctarr}. Research datasets commonly use Neuroimaging Informatics Technology Initiative (NIfTI) volumes, whereas clinical archives retain the Digital Imaging and Communications in Medicine (DICOM) format \cite{li2016dicom}. The raw DICOM can further complicate preparation because a single study may contain multiple acquisitions, reconstructions, localizers, secondary captures, or geometrically invalid series, only some of which are suitable for a given task \cite{brzus2024dicom}. Consequently, a workflow designed for one dataset or analytical objective may not transfer reliably to another.

Existing medical-imaging toolkits can provide deterministic transforms \cite{cardoso2022monai}, and self-configuring systems derive preprocessing parameters from dataset statistics \cite{isensee2021nnu,myronenko2023auto3dseg}. Although these approaches reduce manual effort and improve reproducibility, they commonly assume curated, task-ready volumes or embed routing and preprocessing logic for a specific application. Therefore, adapting them to changes in data organization or task requirements often requires revising rules and parameters manually. These limitations motivate adaptive decision-making conditioned jointly on data characteristics and task requirements. Large language model (LLM)-based agents offer a potential means of realizing such adaptation, as demonstrated in clinical reasoning and tool-using medical workflows \cite{shi2024ehragent,fallahpour2025medrax,yang2026lungnoduleagent}. However, directly allowing an agent to generate code, numerical parameters, or arbitrary workflow structures is unsuitable for medical-image preparation: invalid identifiers, unsupported operations, and unverifiable outputs may silently propagate into downstream analysis \cite{ruan2024toolemu,debenedetti2024agentdojo}.

This tension between contextual adaptability and controlled execution motivates the central question of this work: \emph{Can CT data preparation adapt to input characteristics and downstream requirements while keeping decision space bounded and automatically validating decisions and outputs?} Our key design principle is to separate semantic decision-making from image manipulation. Specifically, a bounded policy proposes a candidate decision specifying \emph{which} eligible input or predefined processing profile should be used, while deterministic components determine whether and \emph{how} it is accepted and executed. This separation introduces adaptability only at decision points that require contextual reasoning, without exposing the underlying workflow to unrestricted generation.

To realize this principle, we propose CT-PrepAgent, which follows a fixed workflow: inspection, bounded semantic decision, guarded resolution, deterministic execution, output verification, and recording. A deterministic inspection characterizes the input data and task as structured profiles. Then, a bounded policy decides an eligible series or a predefined preprocessing profile from the allowed decision space. The controlled execution flow determines whether and how the resulting candidate decision is executed, while the downstream evaluator measures utility only after the decision has been frozen. Linked run records preserve the accepted decision, resolved operation, and verification result for audit and replay. We evaluate the approach on three public CT segmentation tasks and two private raw-DICOM cohorts with registration, together with controlled fault and replay studies. 

Our main contributions are as follows: \noindent
\textbf{(1)} We formulate adaptive CT data preparation as a data-task conditioned decision problem that separates bounded semantic decision from fixed image operations.
\textbf{(2)} We develop a profile-conditioned bounded policy mechanism in which deterministic inspection constructs structured data--task profiles and the policy decides only from eligible series or predefined preprocessing profiles.
\textbf{(3)} We introduce a controlled execution flow that guards, resolves, executes, verifies, and records each decision, supporting bounded recovery, safe quarantine, audit, and policy-free replay. 
\textbf{(4)} We demonstrate competitive segmentation utility, expanded verified series coverage with similar registration metrics on common outputs, and correct fault disposition and drift detection under controlled tests.





\begin{figure*}[t]
\centering
\includegraphics[width=0.95\textwidth]{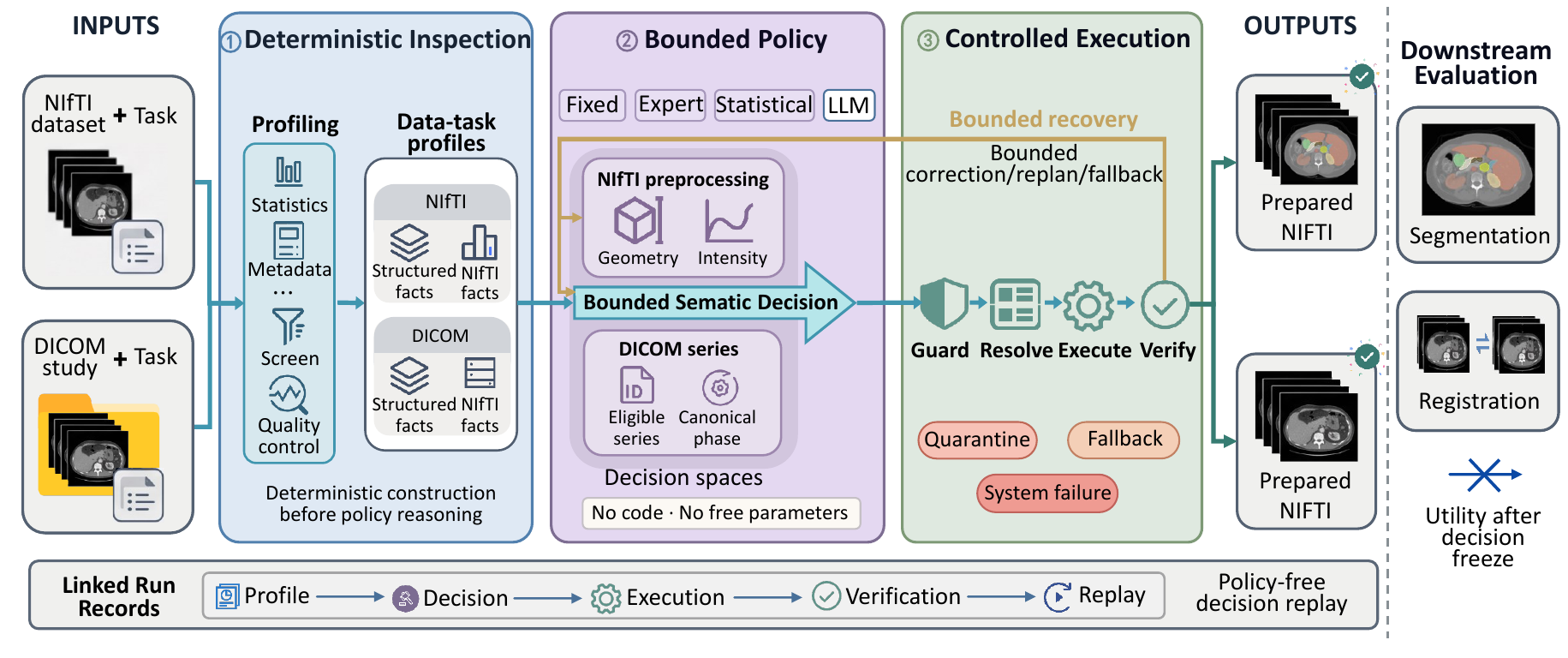}
\caption{Overview of CT-PrepAgent. The deterministic inspection constructs structured data-task profiles from public NIfTI datasets or raw-DICOM studies. A bounded policy decides an eligible series or predefined preprocessing profile. The controlled execution flow guards, resolves, executes, and verifies the candidate decision, while bounded recovery and quarantine contain failures. Linked run records support audit and decision replay; downstream utility is used only for evaluation.}
\label{fig:overview}
\end{figure*}

\section{Methodology}
\label{sec:methodology}
The workflow of CT-PrepAgent, illustrated in Figure~\ref{fig:overview}, transforms CT input $D$ into task-ready output $Y$ for downstream task $T$. It comprises three modules: deterministic inspection, bounded policy, and controlled execution. The \textit{Deterministic Inspection} constructs structured data--task profiles.
The \textit{Bounded Policy} maps an allowlisted view of these profiles to a typed candidate decision: a predefined geometry--intensity pair for NIfTI preparation or an eligible DICOM series with its canonical phase.
Fixed, expert-authored, statistical, and LLM policies operate under the same bounded decision space and cannot generate new operators, numerical parameters, or workflow topology.
The \textit{Controlled Execution} guards the candidate decision, resolves an accepted decision into a versioned deterministic operator, executes, and verifies the resulting output against task-specific postconditions. Only explicitly recoverable outcomes may trigger bounded correction, replan, or fallback; other unusable inputs are quarantined and operator exceptions terminate as system failures.
Finally, linked run records store the profile, accepted decision, resolved operation, verification result, and state transitions needed for audit and policy-free replay.

\subsection{Problem Formulation}
Given CT input \(D\) and downstream task \(T\), the deterministic inspection constructs data and task profiles \((F_D,F_T)=I(D,T)\).
A fixed disclosure map exposes the allowlisted policy view \(F^\pi=H(F_D,F_T)\).
Let \(\mathcal{C}_{D,T}\) denote the finite catalog of executable choices and \(\mathcal{D}_{D,T}\) the data-specific non-execution requests, which may be empty.
The policy decides a typed candidate intended for \(\mathcal{A}_{D,T}=\mathcal{C}_{D,T}\cup\mathcal{D}_{D,T}\):
\begin{equation}
\widetilde C=\pi(F^\pi;\Delta).
\label{eq:policy_proposal}
\end{equation}
Here, \(\Delta\) is empty initially and may contain finite sanitized feedback only when the configured recovery budget permits.
Because a candidate decision may be malformed or inconsistent with the current action space, the guard canonicalizes it or rejects it:
\begin{equation}
g=G(F_D,F_T,\widetilde C)
\in\mathcal{C}_{D,T}\cup\mathcal{D}_{D,T}\cup\{\bot\},
\label{eq:guarded_decision}
\end{equation}
The symbol \(\bot\) denotes pre-execution rejection; a valid \(g\in\mathcal{D}_{D,T}\) enters controlled disposition without image manipulation.
For an accepted executable decision \(g=C\in\mathcal{C}_{D,T}\), the resolver, deterministic operator, and verifier compute:
\begin{equation}
\Omega_C=R(F_D,F_T,C), 
Y_C=O_{\Omega_C}(D),
z=V(Y_C,C,\Gamma_T),
\label{eq:verified_execution}
\end{equation}
where the frozen output specification \(\Gamma_T\) defines verified coverage, geometry, label or series-consistency, and acceptance requirements, and \(z\) is the verifier outcome.
The flow combines \(z\) with any preceding guard, precondition, or execution outcome to determine verified completion, bounded recovery, quarantine, or system failure.

Let \(\mathcal{C}^{\mathrm{ver}}_{D,T}\subseteq\mathcal{C}_{D,T}\) contain the canonical decisions that can produce outputs accepted by the verifier.
When \(\mathcal{C}^{\mathrm{ver}}_{D,T}\neq\emptyset\), the ideal evaluation target is:
\begin{equation}
C^{*}
\in
\arg\max_{C\in\mathcal{C}^{\mathrm{ver}}_{D,T}}
U(Y_C,T),
\label{eq:optimal_decision}
\end{equation}
where \(U\) is task-specific downstream utility, such as segmentation or registration utility.
This definition specifies the target used for evaluation rather than an online optimization procedure: \(U\) is unavailable to the policy and execution flow and is measured only after the decision has been frozen.
If no verified output is available, the execution flow safely quarantines the case rather than selecting an unsafe candidate.

\subsection{Deterministic Inspection for Profiling}
\label{sec:inspector}
The deterministic inspection forms the boundary between heterogeneous CT inputs and policy reasoning.
It converts data-derived observations and task requirements into a compact, reproducible profile.
This design gives every policy the same evidence and keeps source paths, downstream utility, and safety-critical eligibility outside semantic reasoning.

\subsubsection{NIfTI Dataset Profiling}
For a NIfTI dataset containing \(N\) training volumes, the inspection extracts voxel spacing \(s_i\), volume shape \(v_i\), and a whole-volume CT intensity summary \(h_i\) for each case \(i\); the anisotropy ratio is \(a_i=\max(s_i)/\min(s_i)\).
Let \(S=\{s_i\}_{i=1}^N\), \(V=\{v_i\}_{i=1}^N\), and \(A=\{a_i\}_{i=1}^N\).
With target labels \(L_T\) supplied by the task, the data profile is:
\begin{equation}
  F_D^{ni} = (N, Q(S), Q(V), Q_{0.95}(A), h_D, L_T).
\end{equation}
Here, \(Q(\cdot)=(Q_{0.1},Q_{0.5},Q_{0.9})\) gives componentwise quantiles, \(Q_{0.95}(A)\) summarizes upper-tail anisotropy, and \(h_D\) aggregates whole-volume intensity statistics.
The task profile \(F_T^{ni}\) records task type, anatomy, modality, and labels; all data-derived statistics use training cases only, and the test partition is excluded from policy input.

\subsubsection{DICOM Study Profiling}
A raw-DICOM study \(j\) may contain \(K_j\) acquisitions or reconstructions, only some of which satisfy the task requirements.
For each series \(j_k\), the inspection extracts identifier \(u_{j_k}\), de-identified metadata \(m_{j_k}\), geometry \(g_{j_k}\), convertibility \(c_{j_k}\), conversion status \(r_{j_k}\), quality-control result \(q_{j_k}\), and hard-rejection indicator \(h_{j_k}\):
\begin{equation}
  x_{j_k} = (u_{j_k}, m_{j_k}, g_{j_k}, c_{j_k}, r_{j_k}, q_{j_k}, h_{j_k}).
\end{equation}
Before policy invocation, eligibility is determined by
\begin{equation}
\begin{aligned}
e_{j_k}
=
\mathbb{I}\big[
&c_{j_k}
\land (r_{j_k}=\mathrm{available}) \\
&\land
\big(q_{j_k}\in
\{\mathrm{pass},\mathrm{review}\}\big)
\land \neg h_{j_k}
\big].
\end{aligned}
\end{equation}
The resulting eligible identifier set and study profile are
\begin{equation}
\mathcal{E}_j=\{u_{j_k}\mid e_{j_k}=1\},\qquad
F_D^{\mathrm{di}}
=
\left(
K_j,
\{x_{j_k}\}_{k=1}^{K_j},
\mathcal{E}_j
\right).
\end{equation}
Thus, failed conversion, rejected quality control, invalid geometry, and hard-rejected acquisitions are removed before semantic decision.
For DICOM data preparation, \(F_T^{di}\) denotes the task requirement set used by eligibility and resolution.
The policy receives no source paths, and any series it selects must belong to \(\mathcal{E}_j\), separating deterministic eligibility from task-conditioned series choice.

\subsection{Bounded Policy for Semantic Decision}
\label{sec:bounded_policy}
After inspection determines the admissible choices, the policy performs only context-dependent semantic decision space. Fixed, expert-authored, statistical, and LLM-based policies differ only at this stage, while eligibility checks, image operations, and output acceptance criteria remain shared. This bounded decision formulation makes policy effects comparable and prevents semantic reasoning from changing the underlying workflow.

Each policy receives \(F^\pi\) and returns one schema-conforming candidate decision from the current finite action space.
Auxiliary confidence and evidence fields are retained only for audit; they neither expand the action space nor alter the resolved operator. Therefore,
all policy variants differ only in action selection, while sharing the same inspector, guard, resolver, operators, verifier, and evaluator.

\subsubsection{NIfTI Preprocessing Decision}
For NIfTI preparation, the decision space is the Cartesian product of a geometry skill set \(\mathcal{C}_g\) and an intensity skill set \(\mathcal{C}_i\):
\begin{equation}
  \mathcal{C}^{ni} = \mathcal{C}_g \times \mathcal{C}_i,
|\mathcal{C}_g|=2,
|\mathcal{C}_i|=6.
\end{equation}
The geometry skill set contains fixed and planner-derived isotropic resampling, while the intensity skill set contains global \(z\)-score normalization and five anatomy-specific profiles for abdominal soft tissue, abdominal organs, lung, bone, and mediastinum.
Guided by the data--task profiles, the policy decides:
\begin{equation}
  \widetilde C^{ni} = (b_g, b_i), b_g \in \mathcal{C}_{g}, b_i \in \mathcal{C}_{i}.
\end{equation}
A deterministic resolver translates the selected skill pair into executable parameters. Image orientation, label interpolation, cropping, and workflow order remain controlled by the fixed processing contract. Thus, the policy decides a semantic preprocessing profile without directly manipulating numerical parameters.

\subsubsection{DICOM Series Decision}
For one DICOM study \(j\), the policy either decides a series from the eligible set \(\mathcal{E}_j\) or issues a controlled non-execution decision. Its decision space is:
\begin{equation}
\begin{aligned}
  \widetilde{\mathcal{C}}_j^{di}
  ={}&
  \{(\mathrm{use},u,p)\mid u\in\mathcal{E}_j,p\in\mathcal{P}\}\\
  &\cup\{(\mathrm{fallback},\varnothing,\mathrm{unknown})\}\\
  &\cup\{(\mathrm{quarantine},\varnothing,\mathrm{unknown})\},
\end{aligned}
\end{equation}
where $u$ is the decided series identifier and $p$ belongs to the canonical acquisition phase set \(\mathcal{P}\):
\begin{equation}
\begin{aligned}
\mathcal{P}=\{&
\mathrm{noncontrast},
\mathrm{arterial\_or\_ccta},
\mathrm{venous},\\
&
\mathrm{delayed},
\mathrm{unknown}
\}.
\end{aligned}
\end{equation}
The two non-execution decisions instantiate \(\mathcal{D}_{D,T}\) and request a controlled disposition rather than an executable series preprocessing; only a guarded \(\mathrm{use}\) decision enters the resolver.
The policy cannot select a hard-rejected series or determine voxel spacing, cropping, interpolation, or registration parameters. Once a series is accepted, these operations follow the fixed profile associated with the data source and task.
Overall, the bounded policy decides an eligible input, a predefined skill, or a controlled non-execution decision, while the controlled execution flow determines whether and how the candidate decision is accepted and executed.

\subsection{Controlled Execution Flow}
\label{sec:controlled_flow}
A typed candidate decision is not yet an executable plan: it may still be inconsistent with the current input, and its output may fail task requirements.
The controlled execution flow therefore places a guard before image manipulation and a verifier after it, with a deterministic resolver and operator between them.
This two-sided validation keeps the meaning of the policy choice adaptable while making execution and acceptance invariant across policies:
\begin{equation}
\widetilde C
\xrightarrow{G}
C
\xrightarrow{R}
\Omega_C
\xrightarrow{O}
Y_C
\xrightarrow{V}
z.
\end{equation}

\subsubsection{Pre-execution Guard} 
The guard checks schema validity, consistency between the decision and the current profile, and, for executable decisions, membership in the NIfTI decision set or DICOM eligible set.
Only a canonical decision reaches the resolver, which then checks the fixed preconditions of the corresponding operator.
Invalid identifiers, unsupported skills, stale inputs, or inconsistent parameters are therefore rejected before image manipulation.

\subsubsection{Deterministic Execution and Postcondition Verification}
For accepted decision \(C\), the resolver constructs the versioned specification \(\Omega_C\), the fixed operator produces output \(Y_C\), and the verifier evaluates \(Y_C\) against output specification \(\Gamma_T\) as defined in Equation~\ref{eq:verified_execution}.
For NIfTI datasets, the preparation operator materializes geometry and label-preserving transforms and records the selected intensity profile, which the shared downstream loader applies; the verifier checks case coverage, geometry, and label preservation.
For DICOM studies, the operator processes only the accepted series under the frozen cohort profile, and the verifier checks successful materialization, series consistency, spatial geometry, orientation, and spacing.

\subsubsection{Bounded Recovery and Controlled Disposition}
Let \(\rho\) summarize the execution-flow outcome. It equals \(\mathrm{pass}\) only after successful verification, and otherwise records a guard, precondition, execution, or verification failure.
For recoverable and unusable outcome sets \(\mathcal{Z}_{\mathrm{rec}}\) and \(\mathcal{Z}_{\mathrm{un}}\), let \(\mathcal{R}(\rho)\) denotes the configured recovery actions still available for outcome \(\rho\).
The next disposition is
\begin{equation}
d(\rho)=
\begin{cases}
\mathrm{verified},
& \rho=\mathrm{pass},
\\
\mathrm{recover},
& \rho\in\mathcal{Z}_{\mathrm{rec}},
  \ \mathcal{R}(\rho)\neq\emptyset,
\\
\mathrm{quarantine},
& \substack{
\rho\in\mathcal{Z}_{\mathrm{un}}{}\lor\bigl(
\rho\in\mathcal{Z}_{\mathrm{rec}}
\land \mathcal{R}(\rho)=\emptyset
\bigr)
}.
\\
\mathrm{system\ failure},
& \rho=\mathrm{exception}.
\end{cases}
\label{eq:disposition}
\end{equation}
Recovery may correct a malformed decision, select from the remaining eligible actions, or invoke a fixed fallback for a configured failure category. It never expands the decision space or weakens verification.
The execution flow therefore seeks a correct disposition---verified completion or explicit safe termination---rather than completion at any cost.

\subsubsection{Run Records and Decision Replay}
Because verification is decision-specific, the execution flow links the input profile, accepted decision, resolved operation, and verification result for each run.
Decision replay fixes the accepted decision and operator path without reinvoking the policy: input and plan drift are rejected before operator invocation, and the semantic output signature is checked afterward.
Replay thus tests the recorded decision path and output semantics.

\section{Experiments}
\subsection{Experimental Setup}
\subsubsection{Datasets and Implementation}
Our experiments utilize three public NIfTI segmentation datasets with different anatomical and geometric characteristics: AMOS \cite{ji2022amos} (48 training/16 test volumes; 15 abdominal structures), TotalSegmentator Bone (TsBone; 65 training/9 test volumes; six structures), and TotalSegmentator Lung (TsLung; 54 training/15 test volumes; one structure) \cite{wasserthal2023totalsegmentator}. Dataset profiles use training cases only, and all decisions are frozen before downstream training; test cases are not read by the inspection or policy. All policies then use the same 3D U-Net \cite{cicek2016unet}, training protocol, periodic test, and best-checkpoint rule. 

In addition, we use two private de-identified raw DICOM cohorts and adopt 3D CT-to-CT registration as the evaluation: 66 abdominal CT studies and 132 coronary CT angiography (CCTA) studies. Before policy evaluation, a frozen hash-based rule selects 30 studies from each cohort, yielding 60 studies without result-dependent sampling. For each cohort, five volumes with valid paired outputs are evaluated under
two fixed known rigid-transform perturbations, yielding ten registration trials per method per cohort.

\begin{table*}[t]
\centering
\normalsize
\setlength{\tabcolsep}{5pt}
\renewcommand{\arraystretch}{1.08}
\begin{tabular}{@{}llcccc@{}}
\toprule
Dataset & P-D Decided Profile & P-A (Fixed) & P-B (Expert-SOP) & P-C (Statistical) & P-D (Ours) \\
\midrule
AMOS
& Planner 1.1 mm / Abd-soft
& 0.6368 & 0.6905 & 0.6484 & \textbf{0.6987} \\

TsBone
& Planner 1.5 mm / Bone
& \textbf{0.7082} & 0.7021 & 0.7002 & 0.6873 \\

TsLung
& Fixed 1.5 mm / Lung
& 0.8914 & 0.8806 & 0.8960 & \textbf{0.8966} \\
\midrule
 Macro-average
& 
& 0.7455 & 0.7577 & 0.7482 & \textbf{0.7609} \\
\bottomrule
\end{tabular}
\caption{Public NIfTI preparation and segmentation results. ``P-D Decided Profile'' gives P-D's resolved geometry choice, isotropic spacing, and semantic intensity profile. Profiles were computed using training cases only, and P-D decisions were frozen before training. Macro-average is the mean of three dataset-level Dice values. Bold indicates the best result; all P-D decisions were valid and fallback-free.}
\label{tab:nifti_dice}
\end{table*}

\begin{figure*}[t]
\centering
\includegraphics[width=0.98\textwidth]{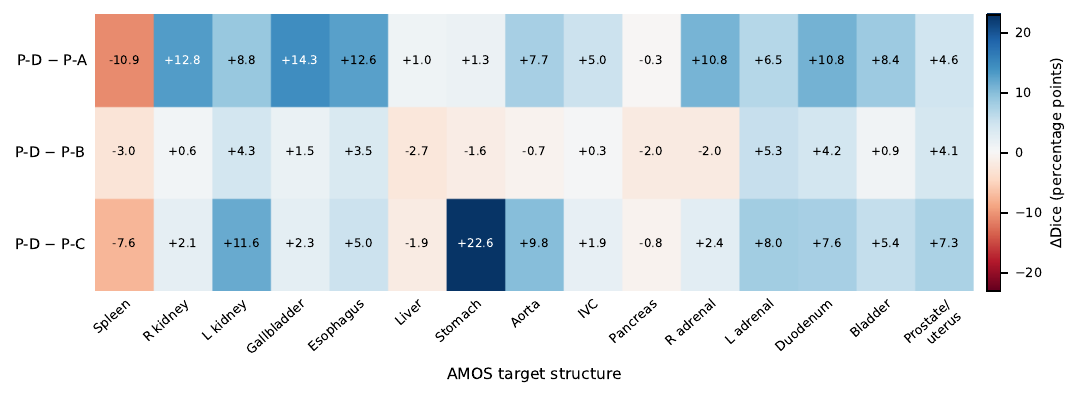}
\caption{AMOS organ-level Dice differences for P-D relative to each baseline under the test protocol. Values are Dice percentage points; blue denotes positive and red negative differences. All 15 target structures are shown without selection.}
\label{fig:amos_delta_dice}
\end{figure*}

\begin{figure*}[t]
    \centering

    \begin{minipage}[t]{0.45\textwidth}
        \centering
        \includegraphics[width=\linewidth]{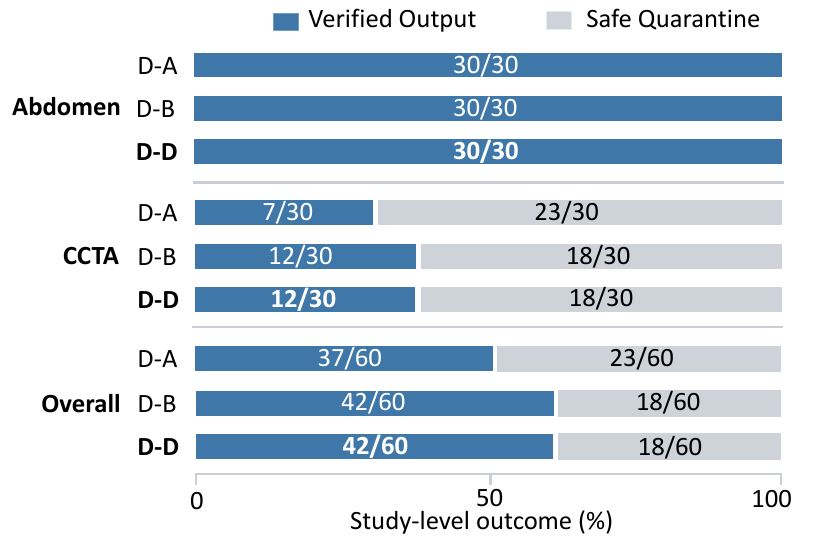}\\
        \textbf{(a) Study-level outcomes}
    \end{minipage}
    \hspace{0.02\textwidth}
    \begin{minipage}[t]{0.45\textwidth}
        \centering
        \includegraphics[width=\linewidth]{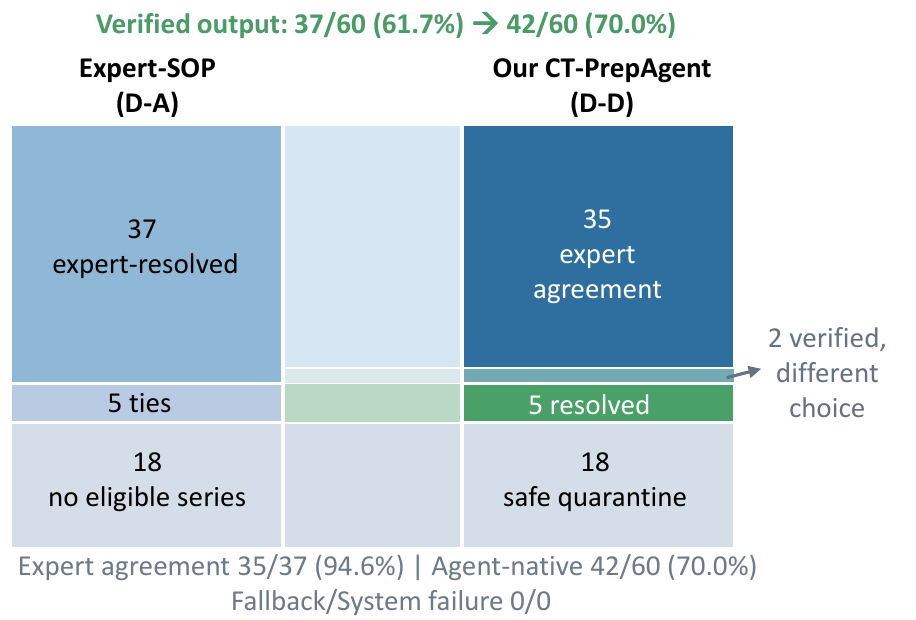}\\
        \textbf{(b) Origin of five additional verfied outputs}
    \end{minipage}

    \caption{Private DICOM preparation results.
    (a) Verified outputs and safe quarantines for the Expert-SOP (D-A), one-shot CT-PrepAgent (D-B), and full CT-PrepAgent (D-D).
    (b) CT-PrepAgent resolves five CCTA studies with tied expert-priority candidates through bounded decision, while all 18 studies without an eligible series remain quarantined. D-B and D-D yield identical nominal results, with all 42 CT-PrepAgent outputs completed without fallback or system failure.}
    \label{fig:dicom_outcomes}
\end{figure*}

\subsubsection{Comparison Methods and Evaluation Metrics}
For public NIfTI preparation, bounded LLM policy P-D is compared with fixed global (P-A), expert-standard operating procedure (SOP) (P-B), and nnU-Net-inspired statistical (P-C) policies \cite{isensee2021nnu} under the same operators and evaluator. We report test Dice and its macro-average Dice.

For private DICOM preparation, the full CT-PrepAgent (D-D) is compared with a frozen expert-SOP (D-A) and a one-shot variant without recovery (D-B); D-C appears only in the recovery ablation. We report verified output yield, expert agreement, and final disposition outcomes, and evaluate registration with median and 95th-percentile target registration errors (TRE50 and TRE95) \cite{fitzpatrick2001distribution}, normalized cross-correlation improvement \(\Delta\)NCC \cite{avants2011reproducible}, and runtime.

For recovery ablation, we compare D-B, D-C, and D-D under identical initial decisions, reporting valid recovery, safe quarantine, correct disposition, incorrect continuations, and unhandled exceptions. For replay robustness, we compare exact replay with controlled input, plan, and semantic drift, reporting correct handling, executor calls, and detection stage under zero LLM calls.

\subsection{Results and Discussion}
We examine whether bounded decision improves downstream utility or usable coverage, and whether controlled execution and replay provide the intended failure containment and reproducibility.
\subsubsection{Results on Public NIfTI Preparation for Segmentation}
As shown in Table~\ref{tab:nifti_dice}, CT-PrepAgent (P-D) produces valid, fallback-free decisions for all three datasets and resolves different preprocessing profiles across them. It obtains Dice values of 0.6987, 0.6873, and 0.8966 on AMOS, TsBone, and TsLung, respectively, yielding the highest macro-average Dice. The clearest gain appears on anisotropic AMOS, where P-D exceeds P-A and P-C by 0.0619 and 0.0503. On TsBone, the fixed policy performs best, suggesting that a simple profile is sufficient for some homogeneous datasets; nevertheless, CT-PrepAgent remains fallback-free and achieves the best Dice on TsLung. These results are consistent with the design goal of selecting among bounded preprocessing profiles according to data and task characteristics rather than enforcing a single universal rule. Overall, CT-PrepAgent adapts CT preparation to heterogeneous data and task requirements while delivering
the strongest aggregate downstream utility.

Figure~\ref{fig:amos_delta_dice} resolves the AMOS result by target structure. P-D exceeds P-A for 13 of 15 structures and is highest among the four policies for 9 of 15. Notable gains relative to P-A appear for the gallbladder, esophagus, adrenal glands, duodenum, and bladder, while the heatmap also retains negative results such as the spleen. The organ-level pattern is consistent with the aggregate AMOS improvement without implying uniform improvement across baselines.



\subsubsection{Results on Private DICOM Preparation for Registration}
Figure \ref{fig:dicom_outcomes} summarizes private DICOM preparation. All methods produce verified outputs for all 30 Abdomen studies. On CCTA, CT-PrepAgent resolves 12/30 studies, compared with 7/30 for expert-SOP. Across both cohorts, CT-PrepAgent increases verified output yield from 37/60 (61.7\%) to 42/60 (70.0\%) and reduces safe quarantines from 23 to 18. The five additional outputs come from CCTA studies with tied expert-priority candidates, which CT-PrepAgent resolves through bounded decision; all 18 studies without an eligible series remain quarantined. CT-PrepAgent agrees with the expert-SOP on 35/37 expert-resolved studies (94.6\%); its two differing selections satisfy the configured postconditions. All 42 outputs are produced without fallback or system failure. D-B and D-D yield the same nominal outcomes because valid initial decisions pass verification and studies without eligible series offer no safe recovery. Thus, the nominal evaluation measures bounded decision coverage, while the controlled-fault study separately assesses recovery and disposition.
The simultaneous gain in coverage and unchanged quarantine of ineligible studies is consistent with the intended separation between deterministic eligibility and policy-based ranking within the eligible set.

Table~\ref{tab:reg_results} evaluates registration utility on paired studies jointly resolved by D-A and D-D. Both methods succeed in all ten trials per cohort. Relative to D-A, D-D changes median TRE95 by only \(+0.0016\) mm on abdomen and \(+0.0175\) mm on CCTA, while maintaining nearly identical \(\Delta\)NCC and comparable runtime. D-D therefore yields similar registration metrics on the common verified outputs.
Because the registration pipeline is fixed, this comparison isolates the effect of series selection on the common subset; it does not assess registration utility for the five outputs.

\begin{table}[t]
\centering
\footnotesize
\setlength{\tabcolsep}{3pt}
\begin{tabular}{@{}llcccc@{}}
  \toprule
  Dataset
  & Method
  & \shortstack{TRE50\\(mm) $\downarrow$}
  & \shortstack{TRE95\\(mm) $\downarrow$}
  & $\Delta$NCC $\uparrow$ 
  & \shortstack{Time\\(min)} \\
  \midrule
  Abdomen
  & D-A (Expert-SOP)
  & 0.0165 & 0.0245 & +0.6130 & 0.1318 \\
  & D-D (Ours)
  & 0.0159 & 0.0261 & +0.6127 & 0.1331 \\
  \midrule
  CCTA
  & D-A (Expert-SOP)
  & 0.0507 & 0.0653 & +0.4698 & 0.0159 \\
  & D-D (Ours)
   & 0.0577 & 0.0828 & +0.4698 & 0.0145 \\
  \bottomrule
\end{tabular}
\caption{Private DIDOM verified output's rigid registration results. Lower TRE and higher $\Delta$NCC indicate better registration. Each method is evaluated over ten trials per cohort.}
\label{tab:reg_results}
\end{table}

\subsubsection{Ablation Study of the Controlled Execution Flow}
\begin{figure}[t]
\centering
\includegraphics[width=0.98\columnwidth]{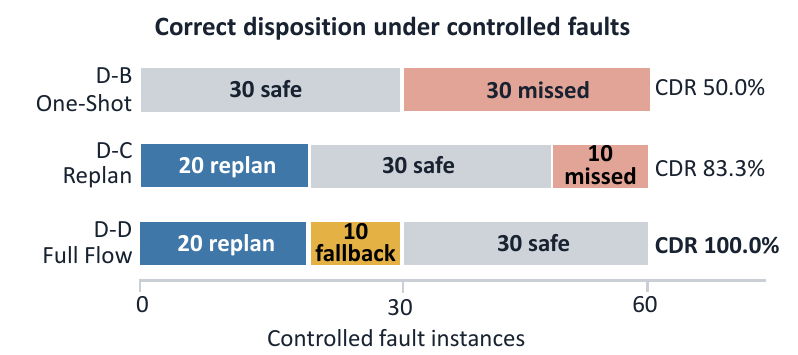}
\caption{Controlled execution ablation on 60 controlled faults.
Bounded replan and expert-SOP fallback improve correct disposition from 50.0\% to 83.3\% and 100.0\%, while all unrecoverable cases are safely quarantined with no incorrect continuation or unhandled exception.}
\label{fig:harness_ablation}
\end{figure}

We evaluate bounded replan and expert-SOP fallback on 60 controlled fault instances balanced between recoverable cases with a valid alternative and unrecoverable cases requiring quarantine. D-B, D-C, and D-D receive identical initial decisions, isolating recovery from initial decision quality. Figure~\ref{fig:harness_ablation} shows that they recover 0/30, 20/30, and 30/30 recoverable cases, yielding correct disposition rates of 50.0\%, 83.3\%, and 100.0\%, respectively. All variants safely quarantine the 30 unrecoverable cases without incorrect continuation or unhandled exceptions. The stepwise gain isolates complementary mechanisms: replan exploits a remaining eligible action, while fallback covers configured recoverable cases not resolved by replan. Unchanged quarantine on unrecoverable faults shows that recovery remains inside the same eligibility boundary.

\subsubsection{Robustness Evaluation of Decision Replay}
We evaluate replay using nine frozen decisions selected before observing replay outcomes: the three public P-D decisions and three accepted D-D decisions from each private cohort. Exact replay checks the input and recorded plan, invokes the frozen operator path without policy calls, and independently checks the semantic signature. Controlled tests introduce input, plan, and semantic drift across the public data and both private cohorts. Table~\ref{tab:replay} shows that all nine exact replays are accepted with zero LLM calls, while all nine controlled mutations are rejected at the intended boundary, yielding 18/18 correct handling. The rejection boundary matches the two-level replay design: content hashes block stale inputs or plans before execution, while semantic signatures detect output drift afterward. Together with acceptance of unchanged paths without policy calls, this separation supports policy-free replay within the controlled mutation model.

\begin{table}[t]
\centering
\footnotesize
\setlength{\tabcolsep}{2.8pt}
\renewcommand{\arraystretch}{1.05}
\begin{tabular}{@{}lrrrl@{}}
\toprule
Condition
& Trials
& \shortstack{Correct\\handling}
& \shortstack{Operator\\calls}
& Outcome \\
\midrule
Exact replay
& 9 & 9/9 & 9 & Accepted \\

Input drift
& 3 & 3/3 & 0 & Rejected before \\

Plan drift
& 3 & 3/3 & 0 & Rejected before \\

Semantic drift
& 3 & 3/3 & 3 & Rejected after \\
\midrule

\textbf{Overall}
& \textbf{18}
& \textbf{18/18}
& \textbf{12}
&  \\
\bottomrule
\end{tabular}
\caption{Robustness of decision replay. All 18 trials use zero
language-model calls. Correct handling denotes accepting exact
replays and rejecting controlled drift. Before and after indicate
rejection before and after operator invocation.}
\label{tab:replay}
\end{table}

\section{Related Work}
\subsubsection{Adaptive CT Data Preparation}
Medical image preparation usually includes spatial standardization, format conversion, intensity normalization, and quality control \cite{trojani2024impact}. MONAI provides reusable deterministic transforms \cite{cardoso2022monai}, and nnU-Net and Auto3DSeg automate segmentation-pipeline configuration through dataset analysis \cite{isensee2021nnu,myronenko2023auto3dseg}. AutoCT and other application-specific CT pipelines further integrate fixed preprocessing with registration, segmentation, or task-specific analysis \cite{bai2023autoct,jin2024preprocessing, butler2025automated}. These systems improve reproducibility and reduce manual effort, but generally assume curated, model-ready volumes or encode processing logic for a given task, requiring rule or parameter revisions when acquisition conditions or analytical objectives change.

Raw clinical DICOM introduces an additional curation problem because a single examination may contain multiple acquisitions, reconstructions, localizers, secondary captures, and series with incompatible geometry or quality. Prior work classifies DICOM series from header attributes or applies task-specific eligibility and priority rules before conversion and downstream analysis \cite{brzus2024dicom,garrett2025methodology}. Imaging infrastructures support large-scale DICOM curation, de-identification, quality review, triggered computation, and provenance capture \cite{denner2023efficient,lemarechal2025paradim,nikolov2025aiready}. However, their routing and preprocessing logic is typically prescribed for a specific task. CT-PrepAgent instead exposes both study-level series selection and dataset-level preprocessing-profile selection through the same bounded policy decision, while eligibility, execution, and output acceptance remain independent of the policy implementation.

\subsubsection{Medical Agentic Workflows}
LLM-based agents have shown promise for decomposing complex tasks, planning intermediate actions, and invoking external tools. In medicine, such agents have focused on clinical reasoning, diagnosis, and information interaction. MDAgents adapts the collaboration structure to task complexity \cite{tang2024medagents}, and MediQ and MedLA investigate active information acquisition and structured medical reasoning \cite{kim2024mdagents,li2024mediq,ma2026medla}. Tool-using systems extend this paradigm beyond dialogue: EHRAgent executes code over electronic health records \cite{shi2024ehragent}, MedRAX invokes specialized tools for chest radiography \cite{fallahpour2025medrax}, and LungNoduleAgent coordinates CT detection, report generation, and diagnosis \cite{yang2026lungnoduleagent}. Recent radiology-agent benchmarks further examine planning and tool use in simulated imaging environments, including interaction with DICOM servers and series-selection tools \cite{zheng2024radabench,maksudov2026abra}. Nevertheless, these systems mainly produce answers, reports, or diagnoses rather than prepared imaging outputs governed by executable acceptance checks.

The closest line of work introduces artifact contracts and deterministic workflow executors for adaptive and reproducible medical image processing \cite{zuo2026artifact}. CT-PrepAgent addresses a different operating point: its policy cannot compose a DAG or create processing modules, but only selects an eligible DICOM series or one profile from an input-dependent finite catalog inside a fixed CT-preparation. This restriction makes fixed, expert-authored, statistical, and LLM policies directly interchangeable under the same execution flow. Our evaluation centers on how preparation decisions affect downstream segmentation and registration, verified yield, fault disposition, and semantic replay; run records support audit and replay but are not the primary claimed novelty. Such restrictions also address broader evidence that unconstrained tool-using agents may execute invalid or unsafe actions \cite{ruan2024toolemu,debenedetti2024agentdojo}.

\section{Conclusion}
We introduce CT-PrepAgent, a bounded framework for adaptive and verifiable CT data preparation. By separating semantic decision from deterministic image execution, CT-PrepAgent allows policies to choose only among eligible DICOM series or predefined preprocessing profiles, avoiding unrestricted workflow generation. Experiments on public CT segmentation tasks and private raw-DICOM cohorts show that CT-PrepAgent improves aggregate segmentation utility, expands verified output yield, preserves registration performance on common verified outputs, and supports bounded recovery, safe quarantine, and policy-free replay. These results suggest that CT-PrepAgent offers a practical agentic paradigm for reliable and auditable CT data preparation.

\bibliography{aaai2027}

\clearpage

\noindent
\begin{minipage}{\columnwidth}
\centering
\large\bfseries
Supplementary Material for\\
CT-PrepAgent: Bounded Policy and Controlled Execution for Adaptive CT Data Preparation
\end{minipage}
\vspace{1em}

\setcounter{section}{0}
\setcounter{subsection}{0}
\setcounter{equation}{0}
\setcounter{table}{0}
\setcounter{figure}{0}

\def\CTPrepAgentCombined{}
\ifdefined\CTPrepAgentCombined
\newcommand{\CTPrepEndSupplement}{}
\else
\documentclass[letterpaper]{article} 
\usepackage[submission]{aaai2027} 
\usepackage[hyphens]{url} 
\usepackage{graphicx} 
\urlstyle{rm} 
\def\UrlFont{\rm} 
\usepackage{natbib} 
\usepackage{caption} 
\usepackage{amsmath}
\usepackage{amssymb}
\usepackage{booktabs}
\usepackage{multirow}
\usepackage{placeins}
\frenchspacing 

\pdfinfo{
/TemplateVersion (2027.1)
}

\setcounter{secnumdepth}{2}

\title{Supplementary Material for\\
CT-PrepAgent: Bounded policy and Controlled Execution for Adaptive CT Data Preparation}
\author{Anonymous Submission}
\affiliations{Anonymous Institution}

\begin{document}

\maketitle
\newcommand{\CTPrepEndSupplement}{\bibliography{aaai2027}\end{document}}
\fi

\section{Additional Method Details}
\label{app:method_details}

This section supplements the profiling, bounded decision, and
controlled execution mechanisms defined in the main paper. It specifies the
information exposed to each policy, the structured contracts connecting framework
modules, the predefined decision space, and the records used for audit and
decision replay.

\subsection{Structured Perception and Policy Contracts}

The inspection converts the complete data and task profiles $(F_D,F_T)$ into an
allowlisted policy view:
\begin{equation}
F^\pi=H(F_D,F_T),
\end{equation}
where $H$ is a fixed disclosure map. The resulting view excludes patient
identifiers, source paths, raw image pixels, test-case identifiers and outcomes,
downstream utility, raw conversion logs, and internal eligibility rules.

For public NIfTI data, $F^\pi$ contains only aggregate geometry, intensity,
label, and task information computed from the training split. Test cases
are excluded from profile construction and intensity calibration, and the
inspection terminates if a test-case identifier is encountered. For a DICOM
study, $F^\pi$ contains de-identified candidate identifiers, allowlisted series
facts, and the deterministically derived eligible set. The policy cannot modify
eligibility or select an identifier outside the current eligible set.

Table~\ref{tab:contract_inventory} summarizes the structured contracts
exchanged among the inspection, policy, controlled execution flow, and replay
path. These contracts separate semantic reasoning from data access and
execution: the policy operates on bounded structured facts, whereas
deterministic components retain control over eligibility, parameter resolution,
image operations, and output acceptance.

\begin{table*}[!t]
\centering
\captionsetup{skip=5pt}
\fontsize{8pt}{8.8pt}\selectfont
\setlength{\tabcolsep}{4pt}
\begin{tabular}{@{}lp{0.29\textwidth}p{0.45\textwidth}@{}}
\toprule
Contract & Principal fields & Purpose and validation boundary \\
\midrule
\texttt{DatasetProfile} &
Dataset ID; training-case membership; case count; spacing, shape, intensity,
anisotropy, labels, and task context &
Training-only public observation; construction rejects test identifiers \\
\texttt{PublicDecision} &
Geometry skill; intensity skill; evidence references; resolved spacing; shared
executor settings &
Schema-conforming public decision and its immutable execution payload \\
\texttt{SeriesCandidateProfile} &
De-identified UID; modality and image type; description; slice count,
thickness, and spacing; body part and phase hints; category and priority;
conversion, quality-control, and exclusion fields &
Per-series DICOM facts from which eligibility is derived deterministically \\
\texttt{PrivateDecision} &
Study/cohort label; eligible candidate records; ranked and selected UIDs;
canonical phase; action; confidence; reason codes; evidence references &
Bounded study-level selection; schema and eligibility are checked before
execution \\
\texttt{RunRecord} &
Profile, decision, execution, and verification references; state transitions;
semantic signature; call counts; terminal disposition &
Links provenance and supports audit and policy-free replay \\
\bottomrule
\end{tabular}
\caption{Core structured contracts used by CT-PrepAgent.}
\label{tab:contract_inventory}
\end{table*}

\subsection{Bounded Policy and Decision Space Design}

After inspection fixes the admissible choices, the policy performs only
context-dependent semantic selection. Public and private policies receive
different structured contexts but follow the same principle: each returns one
typed candidate decision from a finite input-dependent space, while confidence,
rationale, and evidence fields are retained for audit and do not affect
execution.

For public data, one temperature-zero policy call is made per dataset before
downstream training. The stored decision is subsequently consumed without
another model call. For private DICOM data, the policy ranks only
deterministically eligible candidates; sanitized failure feedback is introduced
only during a bounded retry. Table~\ref{tab:policy_contexts} summarizes the two policy contexts and their output schemas.
Tables~\ref{tab:public_prompt_template} and
\ref{tab:private_prompt_template} report the prompt templates used by the
bounded LLM policy. Runtime values are inserted only into the indicated
allowlisted fields; the instructions and output schemas remain fixed.

\begin{table*}[!t]
\centering
\captionsetup{skip=5pt}
\fontsize{8pt}{8.8pt}\selectfont
\setlength{\tabcolsep}{3.5pt}
\begin{tabular}{@{}p{0.62in}p{2.02in}p{2.02in}p{2.02in}@{}}
\toprule
Track & Policy-visible context & Required decision & Fixed boundary \\
\midrule
Public NIfTI &
Case count; spacing and shape summaries; anisotropy; intensity statistics;
label set; task context; planner-resolved spacing preview &
One geometry skill, one intensity skill, and evidence references; confidence
and a brief rationale are optional &
No images, masks, test outcomes, arbitrary spacing, crop, interpolation,
code, commands, or workflow generation \\
Private DICOM &
Study/cohort label; remaining and eligible candidates; de-identified metadata,
geometry, phase hints, conversion state, quality-control fields, and priority;
sanitized retry feedback when applicable &
Ranked UIDs, selected UID, canonical phase, action, confidence, reason codes,
and evidence references &
No pixels, patient identifiers, source paths, or selection outside the current
eligible set; non-execution decisions require a null UID and unknown phase \\
\bottomrule
\end{tabular}
\caption{Policy-visible contexts and typed decision schemas. Public selection
is performed once at the dataset level, whereas private selection is performed
at the study level over a deterministically eligible candidate set.}
\label{tab:policy_contexts}
\end{table*}

\begin{table*}[!t]
\centering
\captionsetup{skip=5pt}
\fontsize{8pt}{8.8pt}\selectfont
\setlength{\tabcolsep}{4pt}
\begin{tabular}{@{}p{1.12in}p{5.72in}@{}}
\toprule
Prompt component & Fixed template content \\
\midrule
Role and objective &
You are a bounded CT preprocessing policy. Using only the supplied training-set
summary and task context, select exactly one geometry skill and one intensity
skill from the provided action lists. Do not propose a new operation or
parameter. \\
Policy input &
Dataset identifier and case count; spacing and shape quantiles; anisotropy;
whole-volume intensity statistics; label set; task type, anatomy, modality, and
target labels; planner-resolved spacing preview; allowed geometry and intensity
actions; evidence-reference allowlist. \\
Selection constraints &
Use no image, mask, test outcome, or downstream utility. Do not emit spacing,
cropping, interpolation, code, commands, or workflow steps. Every evidence
reference must identify a supplied structured fact. \\
Required output &
Return one JSON object containing
geometry\_skill, intensity\_skill, and
evidence\_refs; confidence and a one-sentence
rationale are optional audit fields. No additional keys or text are
permitted. \\
\bottomrule
\end{tabular}
\caption{Public NIfTI preprocessing prompt template. Dataset-specific values
are populated from the training-only policy view, and the returned action is
frozen before downstream training.}
\label{tab:public_prompt_template}
\end{table*}

\begin{table*}[!t]
\centering
\captionsetup{skip=5pt}
\fontsize{8pt}{8.8pt}\selectfont
\setlength{\tabcolsep}{4pt}
\begin{tabular}{@{}p{1.12in}p{5.72in}@{}}
\toprule
Prompt component & Fixed template content \\
\midrule
Role and objective &
You are a bounded DICOM series-selection policy. Rank only the supplied eligible
series and select one series for the stated task, or issue an allowed
non-execution decision when no safe executable selection can be made. \\
Policy input &
De-identified study and cohort labels; remaining and eligible candidate UIDs;
allowlisted metadata, geometry, phase hints, semantic category, priority,
conversion state, and quality-control fields; task requirements; optional
sanitized failure feedback during the single bounded retry. \\
Selection constraints &
A selected UID must belong to the current eligible set and appear in the
ranking. A use decision requires a selected UID and evidence references.
Fallback or quarantine requires a null UID and the phase
unknown. Do not emit patient identifiers, source paths, pixels,
processing parameters, code, commands, or workflow changes. \\
Required output &
Return one JSON object containing ranked\_uids,
selected\_uid, canonical\_phase, action,
confidence, reason\_codes, and
evidence\_refs. The action must be use,
fallback, or quarantine; no additional keys or text are
permitted. \\
\bottomrule
\end{tabular}
\caption{Private DICOM series-selection prompt template. Candidate facts are
de-identified and eligibility is determined before the policy call; retry
feedback contains only sanitized reason codes and the remaining eligible set.}
\label{tab:private_prompt_template}
\end{table*}

\begin{table*}[t]
\centering
\footnotesize
\setlength{\tabcolsep}{4pt}
\begin{tabular}{@{}p{0.50in}p{1.05in}p{2.70in}p{2.07in}@{}}
\toprule
Input & Decision field & Allowlisted actions & Deterministic interpretation \\
\midrule
NIfTI
& Geometry
& \shortstack[l]{fixed\\planner\_isotropic}
& Fixed 1.5-mm isotropic spacing or training-only spacing planning \\
NIfTI
& Intensity
& \shortstack[l]{global\_zscore\\abdomen\_soft\_tissue\\
abdomen\_organ\\lung\\bone\\
mediastinum}
& Versioned, predefined semantic intensity profile \\
DICOM
& Series
& $u\in\mathcal{E}_j$
& Selection of one identifier from the current eligible set \\
DICOM
& Phase
& \shortstack[l]{noncontrast\\arterial\_or\_ccta\\
venous\\delayed\\unknown}
& Canonical acquisition-phase label \\
DICOM
& \mbox{Non-execution}
& fallback,quarantine
& Controlled disposition without executable series preparation \\
\bottomrule
\end{tabular}
\caption{Complete policy-facing action catalog. The public decision space
contains $2\times6=12$ geometry--intensity combinations. DICOM choices depend
on the current study because the selected UID must belong to $\mathcal E_j$.}
\label{tab:action_catalog}
\end{table*}

The executable choices are drawn from a predefined decision space before formal evaluation. Table~\ref{tab:action_catalog} lists all allowable decisions and their deterministic interpretations.
The fixed geometry skill resolves to 1.5-mm isotropic spacing. For
planner-derived geometry, the resolver begins with the median training
spacing along each axis. If the coarsest median axis exceeds three times the
larger of the other two medians, that axis is replaced by its training-set 10th percentile. The resolver then takes the geometric mean, rounds it to 0.1 mm, and applies the frozen anatomy guard. Both geometry skills use the same deterministic body crop, LPS orientation, one-hot label resampling, and patch size derived from a 144-mm physical field of view. These settings are controlled by the resolver and are not policy actions.
The five anatomy-specific intensity profiles are clipped piecewise-linear HU
mappings:
\begin{equation}
\begin{aligned}
\text{Abdomen\_soft\_tissue}:&\quad (-100,0)\rightarrow(200,1),\\
\text{Abdomen\_organ}:&\quad (-1000,0),(-200,0.2),\\
&\quad (100,0.7),(250,1),\\
\text{Lung}:&\quad (-1000,0)\rightarrow(-200,1),\\
\text{Bone}:&\quad (200,0)\rightarrow(1500,1),\\
\text{Mediastinum}:&\quad (-200,0)\rightarrow(350,1).
\end{aligned}
\end{equation}
The policy selects a profile whose parameters are deterministically resolved and recorded for audit and replay.
After semantic mapping, calibration parameters are computed from sampled
foreground training voxels and frozen. The global\_zscore profile
instead derives its 0.5th- and 99.5th-percentile clipping bounds, mean, and
standard deviation from sampled raw foreground HU values in the training split.
These frozen statistics are applied unchanged to training and test images;
test labels are never used for calibration.
Overall, the bounded policy determines which eligible input or predefined skill should be used. It cannot create operators, generate numerical parameters, alter eligibility, or modify the workflow topology.

\begin{table*}[t]
\centering
\small
\setlength{\tabcolsep}{5pt}
\begin{tabular}{@{}lll@{}}
\toprule
Stage & Required checks or operation & Recorded result \\
\midrule
Guard
& Schema validity, profile consistency, action membership, and input freshness
& Canonical decision or reason code \\
Resolve
& Operator version, resolved parameters, input digest, and fixed preconditions
& Versioned operator specification and plan digest \\
Execute
& Deterministic operator invocation in an isolated run namespace
& Output references and execution summary \\
Verify
& Task-specific geometry, label, coverage, and readiness postconditions
& Verification outcome and semantic signature \\
Recover
& Bounded correction, replan, configured fallback, or quarantine
& Recovery path and terminal disposition \\
\bottomrule
\end{tabular}
\caption{Controlled execution records. An output is accepted only after all
required postconditions pass.}
\label{tab:harness_contract}
\end{table*}

\begin{table}[t]
\centering
\small
\setlength{\tabcolsep}{4pt}
\begin{tabular}{@{}p{0.95in}p{2.05in}@{}}
\toprule
Record & Principal contents \\
\midrule
Profile record & Structured facts, task context, and input digest \\
Decision record & Raw policy output, canonical decision, and evidence \\
Execution record & Operator version, plan digest, and output summary \\
Verification record & Outcome, reason codes, and semantic signature \\
Run-level record & State transitions, record references, call counts, and
terminal state \\
\bottomrule
\end{tabular}
\caption{Typed run records used for audit and decision replay.}
\label{tab:run_record_fields}
\end{table}

\subsection{Controlled Execution and Decision Replay}

Section 2.4 of the main paper defines the Guard--Resolve--Execute--Verify flow. Table~\ref{tab:harness_contract} supplements this definition by specifying the
checks and records produced at each stage.
Each run terminates in one of four states: complete\_agent,
complete\_fallback, quarantined, or
failed\_system. These states denote verified policy completion,
verified fallback completion, the absence of a safe executable path, and
software or I/O failure, respectively.

In the full private flow, a schema-invalid decision permits at most one
correction, and a recoverable candidate-level failure permits at most one
replan over the remaining eligible set. Expert-SOP fallback is available only
for configured recoverable categories after these budgets are exhausted.
Recovery never expands the eligible decision space or weakens output
verification.

Each run is represented by the linked record chain:
\begin{equation}
R_P\rightarrow R_D\rightarrow R_E\rightarrow R_V,
\end{equation}
where $R_P$, $R_D$, $R_E$, and $R_V$ denote the profile, decision, execution,
and verification records, respectively. Every record contains version
information, input references, and a payload digest. A run-level record links
these components with state transitions, policy-call counts, and the terminal
state.

Decision replay loads a frozen accepted decision, verifies the input and plan
digests, and invokes the recorded deterministic operator path in a new namespace
without reinvoking the policy. The verifier then compares the terminal state
and semantic signature with those of the source run. Input and plan drift are
rejected before operator invocation, whereas semantic drift is detected after
execution.

Replay therefore evaluates reproducibility at the level of the validated
decision path and output semantics. It does not require compressed
medical-image files to be byte-identical across runs.

\begin{table*}[t]
\centering
\scriptsize
\setlength{\tabcolsep}{5pt}
\begin{tabular}{@{}llp{0.48\textwidth}@{}}
\toprule
Condition & Implemented method & Information and interpretation \\
\midrule
Fixed/task-agnostic & P-A & Fixed 1.5-mm geometry plus global z-score;
uses neither profile statistics nor anatomy to choose an action. \\
Task/anatomy rule & P-B & Frozen expert task-card mapping to a predefined
profile; closest implemented task-only/anatomy-only control. It is not an
LLM task-only ablation. \\
Data/statistics only & P-C & Planner geometry from training-only spacing
statistics plus global z-score. The selector does not configure or inherit
nnU-Net architecture/training; the same downstream U-Net trainer is held
common across Policies. \\
Full data--task profile & P-D & One bounded LLM choice using the structured
profile and task context, with the same 12-action catalog. \\
Private metadata heuristic & D-A (Expert-SOP) & Frozen eligibility and
metadata ordering; no LLM call. \\
Private one-shot & D-B & Same bounded Agent proposal without correction,
replan, or fallback. \\
Private replan & D-C & Bounded correction/replan but no Expert-SOP fallback. \\
Private full flow & D-D & Bounded correction and replan plus configured
Expert-SOP fallback. \\
Tie-break controls & Random/canonical & Uniform or lexicographic selection
inside the exact unresolved eligible winner set; no LLM call. \\
\bottomrule
\end{tabular}
\caption{Meaning of every principal policy and control. All comparisons
share deterministic eligibility, resolution, execution, and verification
within their track.}
\label{tab:baseline_definitions}
\end{table*}

\section{Experimental Protocol and Reproducibility}

\subsection{Comparison Methods}
We compare CT-PrepAgent with fixed, expert-SOP, and statistical policies for public NIfTI preparation, and with a frozen expert-SOP and controlled-execution variants for private DICOM preparation. Within each setting, all methods share the same deterministic components and downstream evaluator, isolating decision selection and recovery. Table~\ref{tab:baseline_definitions} summarizes all comparison methods and controls.


\begin{table*}[p]
\centering
\small
\setlength{\tabcolsep}{5pt}
\begin{tabular}{@{}lrrrr@{}}
\toprule
Dataset
& P-D Dice
& $\Delta$ vs. P-A [95\% CI]
& $\Delta$ vs. P-B [95\% CI]
& $\Delta$ vs. P-C [95\% CI] \\
\midrule
AMOS
& 0.6987
& $+0.0619\ [ +0.0477,+0.0765]$
& $+0.0082\ [ -0.0012,+0.0172]$
& $+0.0503\ [ +0.0396,+0.0620]$ \\
TsBone
& 0.6873
& $-0.0209\ [ -0.0672,+0.0169]$
& $-0.0148\ [ -0.0304,-0.0032]$
& $-0.0129\ [ -0.0449,+0.0119]$ \\
TsLung
& 0.8966
& $+0.0052\ [ -0.0047,+0.0188]$
& $+0.0160\ [ -0.0004,+0.0406]$
& $+0.0006\ [ -0.0047,+0.0068]$ \\
\bottomrule
\end{tabular}
\caption{Paired Dice differences between CT-PrepAgent (P-D) and public
baselines. AMOS and TsLung satisfy all three predefined descriptive criteria;
TsBone does not.}
\label{tab:paired_ci}

\vspace{1.2em}

\centering
\footnotesize
\setlength{\tabcolsep}{4pt}
\begin{tabular}{@{}p{0.55in}p{0.78in}p{2.95in}p{2.05in}@{}}
\toprule
Dataset & Equivalent methods & Resolved preparation & Interpretation \\
\midrule
AMOS & None & Four distinct configurations
& Numerical differences include preparation differences \\
TsBone & P-A $=$ P-C & 1.5 mm + global $z$-score
& Dice difference 0.0080 reflects evaluator variability \\
TsBone & P-B $=$ P-D & 1.5 mm + bone profile
& Dice difference 0.0148 is not attributed to policy choice \\
TsLung & P-A $=$ P-C & 1.5 mm + global $z$-score
& Dice difference 0.0046 reflects evaluator variability \\
TsLung & P-B $=$ P-D & 1.5 mm + lung profile
& Dice difference 0.0160 is not attributed to policy choice \\
\bottomrule
\end{tabular}
\caption{Resolved-configuration equivalence audit. Equality denotes an
identical content-addressed preprocessing key and a shared prepared payload.}
\label{tab:key_audit}

\vspace{1.2em}

\centering
\scriptsize
\setlength{\tabcolsep}{4.5pt}
\renewcommand{\arraystretch}{1.02}
\begin{tabular}{@{}llrcccc@{}}
\toprule
Dataset & Target structure & P-A & P-B & P-C & P-D \\
\midrule
AMOS & Spleen & \textbf{0.7711} & 0.6922 & 0.7376 & 0.6621 \\
AMOS & Right kidney & 0.7048 & 0.8267 & 0.8121 & \textbf{0.8327} \\
AMOS & Left kidney & 0.6485 & 0.6936 & 0.6207 & \textbf{0.7364} \\
AMOS & Gallbladder & 0.6087 & 0.7369 & 0.7292 & \textbf{0.7519} \\
AMOS & Esophagus  & 0.5535 & 0.6444 & 0.6291 & \textbf{0.6792} \\
AMOS & Liver & 0.8861 & \textbf{0.9232} & 0.9153 & 0.8964 \\
AMOS & Stomach & 0.6871 & \textbf{0.7162} & 0.4741 & 0.6998 \\
AMOS & Aorta & 0.7705 & \textbf{0.8550} & 0.7496 & 0.8476 \\
AMOS & Inferior vena cava & 0.7696 & 0.8170 & 0.8004 & \textbf{0.8197} \\
AMOS & Pancreas & 0.6401 & \textbf{0.6566} & 0.6451 & 0.6367 \\
AMOS & Right adrenal gland & 0.5066 & \textbf{0.6344} & 0.5901 & 0.6143 \\
AMOS & Left adrenal gland & 0.4439 & 0.4561 & 0.4291 & \textbf{0.5088} \\
AMOS & Duodenum & 0.4704 & 0.5370 & 0.5032 & \textbf{0.5787} \\
AMOS & Bladder & 0.5519 & 0.6270 & 0.5815 & \textbf{0.6358} \\
AMOS & Prostate/uterus & 0.5437 & 0.5487 & 0.5158 & \textbf{0.5892} \\
\midrule
TsBone & Left femur & 0.3910 & 0.4976 & \textbf{0.5040} & 0.4394 \\
TsBone & Right femur & \textbf{0.5010} & 0.3945 & 0.4075 & 0.4654 \\
TsBone & Left hip & 0.3877 & \textbf{0.5078} & 0.4547 & 0.4374 \\
TsBone & Right hip & 0.4268 & 0.3595 & 0.3596 & \textbf{0.4275} \\
TsBone & Sacrum & 0.7577 & \textbf{0.8490} & 0.7843 & 0.8183 \\
TsBone & Vertebrae & \textbf{0.9051} & 0.8804 & 0.8909 & 0.8643 \\
\midrule
TsLung & Lung & 0.8914 & 0.8806 & 0.8960 & \textbf{0.8966} \\
\bottomrule
\end{tabular}
\caption{Per-structure Dice over all 22 target structures. Bold denotes the numerical
maximum among the four policies, without implying statistical significance or
a policy effect when the prepared payload is shared.}
\label{tab:per_structure_dice}
\end{table*}


\begin{table*}[t]
\centering
\footnotesize
\setlength{\tabcolsep}{3.5pt}
\begin{tabular}{@{}llrrrrrrr@{}}
\toprule
Cohort & Method & Studies & Verified output & Expert agreement
& Agent-native & Fallback & Quarantine & System failure \\
\midrule
\multirow{3}{*}{Abdomen}
& D-A & 30 & 30 (100.0\%) & -- & -- & 0 & 0 & 0 \\
& D-B & 30 & 30 (100.0\%) & 28/30 (93.3\%) & 30 (100.0\%) & 0 & 0 & 0 \\
& D-D & 30 & 30 (100.0\%) & 28/30 (93.3\%) & 30 (100.0\%) & 0 & 0 & 0 \\
\midrule
\multirow{3}{*}{CCTA}
& D-A & 30 & 7 (23.3\%) & -- & -- & 0 & 23 & 0 \\
& D-B & 30 & 12 (40.0\%) & 7/7 (100.0\%) & 12 (40.0\%) & 0 & 18 & 0 \\
& D-D & 30 & 12 (40.0\%) & 7/7 (100.0\%) & 12 (40.0\%) & 0 & 18 & 0 \\
\midrule
\multirow{3}{*}{Overall}
& D-A & 60 & 37 (61.7\%) & -- & -- & 0 & 23 & 0 \\
& D-B & 60 & 42 (70.0\%) & 35/37 (94.6\%) & 42 (70.0\%) & 0 & 18 & 0 \\
& D-D & 60 & 42 (70.0\%) & 35/37 (94.6\%) & 42 (70.0\%) & 0 & 18 & 0 \\
\bottomrule
\end{tabular}
\caption{Complete private DICOM governance results. D-A is the frozen
Expert-SOP, D-B is one-shot CT-PrepAgent, and D-D is CT-PrepAgent with full
controlled execution.}
\label{tab:private_governance}
\end{table*}

\subsection{Public NIfTI Segmentation Protocol}
\paragraph{Training Protocol.}
We evaluate public NIfTI preparation on AMOS \cite{ji2022amos} and the TsBone
and TsLung tasks derived from TotalSegmentator
\cite{wasserthal2023totalsegmentator} under a shared downstream training
protocol. All policies use the same four-level 3-D U-Net
\cite{cicek2016unet} with 32 base channels, batch size 1, and gradient
accumulation over two steps. AMOS is trained for 300 epochs, while TsBone
and TsLung are trained for 200 epochs, with 25 optimization steps per epoch,
Dice--cross-entropy loss, AdamW \cite{loshchilov2019decoupled}, five warmup
epochs, and a cosine schedule \cite{loshchilov2017sgdr}. The initial learning
rate and weight decay are $2\times10^{-4}$ and $10^{-5}$, respectively. Two
patches are sampled per volume using the same balanced sampler. Test is performed every
25 epochs for AMOS and every 20 epochs for TsBone and TsLung using
sliding-window inference with an overlap of 0.25. The five TsLung lobe labels are merged
into a single foreground class, and each condition uses one training run with
seed 42.
For test case \(i\) and each foreground class \(c\), Dice is defined as:
\[
\operatorname{Dice}_{ic}
=
\frac{2|P_{ic}\cap G_{ic}|}{|P_{ic}|+|G_{ic}|},
\]
where \(P_{ic}\) and \(G_{ic}\) denote the predicted and ground-truth regions,
respectively. We average Dice over foreground classes within each
case and then over test cases to obtain the dataset-level score. The
three-dataset macro-average is the unweighted mean of the dataset-level scores.

\subsection{Private DICOM Registration Protocol}
The expert-SOP ranks eligible series by priority, slice count, slice thickness, and in-plane voxel area. Ties equivalent on all observed acquisition fields are canonicalized, whereas unresolved non-equivalent ties are quarantined. Non-CT acquisitions, localizers, dose reports, screen captures, failed conversions, rejected quality-control results, and invalid 3-D volumes are excluded before policy invocation.

Downstream performance is evaluated by known-transform rigid registration. For
each cohort, five volumes with paired verified outputs are subjected to two
fixed perturbations, yielding ten trials per method. Using base seed
20260713, each perturbation applies per-axis translations within
\(\pm20\) mm and rotations within \(\pm10^\circ\). TRE50 and TRE95 are
computed from 1,000 physical points sampled within a body mask obtained by
thresholding at \(HU>-500\), retaining the largest connected component, and
applying spherical closing with a two-voxel radius. D-A and D-D use identical
perturbations. We also report
\(\Delta\mathrm{NCC}=\mathrm{NCC}_{\mathrm{after}}-
\mathrm{NCC}_{\mathrm{before}}\) and runtime.

\subsection{Implementation Details}
The experiments use PyTorch \cite{paszke2019pytorch} for segmentation and SimpleITK and
ANTs-based components \cite{avants2011reproducible} for volume preparation and registration. The bounded LLM
policy uses \path{Qwen3-Next-80B-A3B-Instruct-6bit} at
temperature zero. It receives only
the allowlisted structured profile and returns the fixed typed decision schema. 

\section{Extended Public Segmentation Results}

\subsection{Case-Level Paired Differences}
We pair policies by test-case ID and bootstrap case-level Dice
differences 2,000 times with replacement. Table~\ref{tab:paired_ci}
reports percentile 95\% confidence intervals for P-D against each
baseline. The prespecified criteria require a positive mean
difference from P-A and lower confidence bounds no smaller than \(-0.03\)
relative to P-B and P-C. These intervals quantify validation-case uncertainty for fixed checkpoints, not training variability. For policy pairs
sharing the same prepared payload, numerical differences reflect downstream
training variability rather than preprocessing selection.

AMOS provides the clearest evidence for adaptive preprocessing. P-D selects
planner\_isotropic and abdomen\_soft\_tissue, resolves to
1.1-mm spacing, and exceeds P-A and P-C by 0.0619 and 0.0503;
its difference from P-B remains uncertain. TsBone yields mixed results. On
TsBone and TsLung, some policy pairs share the same prepared payload, so
their differences cannot be attributed to preprocessing selection. The
TsLung intervals include zero, indicating comparable case-level performance.
Overall, the results support data--task adaptation on heterogeneous inputs
without implying uniform gains across datasets or policies.

\subsection{Preprocessing Equivalence and Attribution}
Two policy decisions are preprocessing-equivalent when they resolve to the
same spacing, intensity transformation, cropping, reorientation, label
interpolation, and executor version, producing an identical prepared NIfTI
payload. Performance differences between equivalent decisions cannot be
attributed to preprocessing selection and instead reflect downstream training
variation. As shown in Table~\ref{tab:key_audit}, equivalent
policy pairs occur on the near-isotropic TsBone and TsLung datasets, whereas
all four policies resolve to distinct configurations on anisotropic AMOS.
AMOS therefore provides the clearest basis for attributing performance
differences to preprocessing selection.



\subsection{Per-Structure Dice}
Table~\ref{tab:per_structure_dice} reports per-structure Dice across the three
public tasks. On AMOS, P-D achieves the best result for 9 of 15 organs and
outperforms P-A for 13 organs, with notable gains for the gallbladder,
esophagus, adrenal glands, duodenum, and bladder. This broad organ-level
pattern supports the aggregate benefit of adaptive preprocessing on
anisotropic abdominal CT, although performance does not improve uniformly
across all structures. TsBone shows mixed structure-level results, indicating
that no preprocessing profile is universally advantageous. TsLung contains
only one whole-lung target, for which P-D achieves the best Dice.

\section{Extended Private DICOM Results}
\label{app:private_selection}

\subsection{Study-Level DICOM Outcomes}
Table~\ref{tab:private_governance} reports the study-level DICOM outcomes,
while Table~\ref{tab:ccta_decomposition} decomposes the CCTA outcomes by
candidate availability and tie status. CT-PrepAgent produces five additional
verified outputs by resolving ties among eligible candidates, while all 18
studies without an eligible series remain quarantined. To determine whether
this gain is LLM-specific, we compare random and canonical tie-breaking after
the frozen Expert-SOP ranks candidates by priority, slice count, slice
thickness, and in-plane voxel area. Both controls produce the same five
additional outputs, and canonical tie-breaking matches all five CT-PrepAgent
selections. Thus, the coverage gain arises from resolving eligible ties rather
than LLM-specific reasoning. CT-PrepAgent also agrees with Expert-SOP on the
UID and phase for 35 of 37 expert-resolved studies. Because independent
intended-series and phase annotations are unavailable, these results do not
establish clinical selection accuracy.

\begin{table}[t]
\centering
\small
\setlength{\tabcolsep}{4pt}
\begin{tabular}{@{}lrrr@{}}
\toprule
CCTA condition & $N$ & D-A & D-B/D-D \\
\midrule
Unique highest priority & 7 & Output & Output \\
Highest-priority tie & 5 & Quarantine & Verified output \\
No eligible candidate & 18 & Quarantine & Quarantine \\
\bottomrule
\end{tabular}
\caption{CCTA decision decomposition. The gain from 7 to 12 verified outputs
comes entirely from bounded resolution of five expert-priority ties.}
\label{tab:ccta_decomposition}
\end{table}

\begin{table*}[!t]
\centering
\captionsetup{skip=3pt}
\scriptsize
\setlength{\tabcolsep}{5pt}
\begin{tabular}{@{}lp{0.33\textwidth}p{0.35\textwidth}@{}}
\toprule
Injected category & Recoverable case (5 total across cohorts) &
Unrecoverable case (5 total across cohorts) \\
\midrule
Missing critical tag & Valid Expert alternative; expected Expert fallback
under D-D & No valid alternative; expected quarantine \\
Modality mismatch & Remaining eligible CT candidate; expected replan &
No eligible CT alternative; expected quarantine \\
Excluded candidate & Valid Expert alternative; expected Expert fallback
under D-D & All candidates excluded; expected quarantine \\
Out-of-candidate UID & Remaining eligible in-profile UID; expected replan &
No eligible in-profile alternative; expected quarantine \\
Conversion failure & A remaining candidate converts; expected replan &
All eligible conversions fail; expected quarantine \\
Invalid spacing/hard QC & A remaining candidate passes checks; expected
replan & No candidate passes geometry/QC; expected quarantine \\
\bottomrule
\end{tabular}
\caption{Complete controlled-fault taxonomy and intended dispositions.
Counts sum to ten instances per row and 60 overall.}
\label{tab:fault_taxonomy}

\vspace{5pt}
\centering
\footnotesize
\textbf{(a) Aggregate disposition results}\par\vspace{2pt}
\setlength{\tabcolsep}{8pt}
\begin{tabular}{@{}lrrr@{}}
\toprule
Method & Valid recovery & Safe quarantine & Correct disposition \\
\midrule
D-B & 0/30 (0.0\%) & 30/30 (100.0\%) & 30/60 (50.0\%) \\
D-C & 20/30 (66.7\%) & 30/30 (100.0\%) & 50/60 (83.3\%) \\
D-D & 30/30 (100.0\%) & 30/30 (100.0\%) & 60/60 (100.0\%) \\
\bottomrule
\end{tabular}

\par\vspace{3pt}
\textbf{(b) Recovery by fault category}\par\vspace{2pt}
\setlength{\tabcolsep}{5pt}
\begin{tabular}{@{}lrrrrl@{}}
\toprule
Recoverable fault & $N$ & D-B & D-C & D-D & D-D recovery path \\
\midrule
Conversion failure & 5 & 0/5 & 5/5 & 5/5 & Replan \\
Excluded candidate & 5 & 0/5 & 0/5 & 5/5 & Expert-SOP fallback \\
Invalid spacing or quality control & 5 & 0/5 & 5/5 & 5/5 & Replan \\
Missing critical metadata & 5 & 0/5 & 0/5 & 5/5 & Expert-SOP fallback \\
Modality mismatch & 5 & 0/5 & 5/5 & 5/5 & Replan \\
Out-of-candidate identifier & 5 & 0/5 & 5/5 & 5/5 & Replan \\
\bottomrule
\end{tabular}
\caption{Controlled-fault disposition and recovery. (a) Aggregate outcomes.
(b) Recovery by failure category. Each method is evaluated on 60 faults: 30
recoverable and 30 unrecoverable. All methods safely quarantine the 30
unrecoverable cases, with no incorrect continuation or unhandled exception.}
\label{tab:fault_results}

\vspace{5pt}
\centering
\footnotesize
\setlength{\tabcolsep}{4pt}
\begin{tabular}{@{}p{0.75in}p{2.45in}p{1.25in}p{1.75in}@{}}
\toprule
Condition & Controlled change & Expected boundary & Expected result \\
\midrule
Exact replay & None & Input, plan, and post-execution checks
& Invoke path and accept \\
Input drift & Current input differs from the frozen input digest
& Before execution & Reject without executor call \\
Plan drift & Operator version or plan digest differs
& Before execution & Reject without executor call \\
Semantic drift & Semantic signature differs after execution
& After execution & Reject after executor call \\
\bottomrule
\end{tabular}
\caption{Decision-replay conditions and expected verification boundaries.
Exact replay preserves the recorded artifacts, whereas the three negative
controls independently introduce input, plan, or semantic drift. All
conditions use zero policy calls.}
\label{tab:replay_evaluation}
\end{table*}

\section{Controlled-Fault Evaluation}
\label{app:fault_replay}

\subsection{Fault Design and Evaluation Criteria}
We construct 60 deterministic fault instances across six categories: missing
critical metadata, modality mismatch, excluded candidates, out-of-candidate
identifiers, conversion failure, and invalid spacing or quality-control
failure. Each cohort contributes five instances per category, yielding
\(2\times6\times5=60\) instances. Across both cohorts, each category contains
five recoverable cases with a known safe alternative and five unrecoverable
cases requiring quarantine, for 30 cases of each type overall.
Table~\ref{tab:fault_taxonomy} summarizes the injected mutation,
recoverability criterion, and expected disposition for each category. 

D-B, D-C, and D-D receive identical scripted initial decisions, isolating the
effects of recovery from initial decision quality. Valid recovery is the proportion of recoverable cases completed using a
verified safe alternative. Safe quarantine is the proportion of
unrecoverable cases correctly quarantined without incorrect continuation or
an unhandled exception. Correct disposition counts valid recovery for
recoverable cases and safe quarantine for unrecoverable cases. These tests
evaluate the configured recovery and termination mechanisms, not real-world
fault prevalence or clinical safety.

\subsection{Recovery and Disposition Results}
Table~\ref{tab:fault_results}(b) decomposes D-D recovery by fault category
and mechanism. Bounded replanning recovers all conversion-failure, invalid
spacing or quality-control, modality-mismatch, and out-of-candidate cases,
yielding \(20/20\) recoveries across these four categories. Expert-SOP
fallback recovers the remaining excluded-candidate and missing-critical-
metadata cases, yielding another \(10/10\) recoveries. Thus, replanning
handles failures with a remaining eligible action, while fallback covers
configured cases that replanning cannot resolve.

As summarized in Table~\ref{tab:fault_results}(a), these complementary
mechanisms increase valid recovery from \(0/30\) for D-B to \(20/30\) for
D-C and \(30/30\) for D-D. All variants safely quarantine the 30
unrecoverable cases without incorrect continuation or unhandled exceptions.
The results show that replanning and Expert-SOP fallback expand recovery
coverage while preserving the same eligibility and verification boundaries.

\section{Decision-Replay Robustness}

\subsection{Replay Design and Verification Boundaries}
To evaluate policy-free replay and drift detection, we select nine frozen
accepted decisions before observing replay outcomes: the P-D decisions for
AMOS, TsBone, and TsLung, and three accepted D-D decisions from each private
cohort. Exact replay preserves the input digest, recorded plan, and semantic
signature. Three negative controls independently introduce input, plan, or
semantic drift, each using one public, one Abdomen, and one CCTA decision.
The evaluation therefore contains nine exact replays and nine controlled
mutations, all with zero policy and LLM calls.
Table~\ref{tab:replay_evaluation} summarizes the controlled changes and
their expected detection boundaries.

Replay checks the input and plan before invoking the recorded operator path
and verifies the semantic signature afterward. Public signatures encode case
and label coverage, resolved spacing, and affine validity; private signatures
encode the selected UID and phase, output shape, spacing, affine, orientation,
and registration readiness. Public replay may reuse a validated
content-addressed payload after key and lock validation, whereas private
replay materializes a new output. Acceptance requires both the expected
terminal state and a matching semantic signature.

\subsection{Replay and Drift-Detection Results}
All nine exact replays invoke the frozen operator path
and pass verification without policy calls. All three input mutations and
three plan mutations are rejected before operator invocation, preventing six
executor calls, while all three semantic mutations are detected after
execution. Overall, the system accepts all nine unchanged decisions and
rejects all nine controlled mutations, yielding \(18/18\) correct handling.
The quarantined and failed\_system states produced by the
negative controls are expected drift-detection outcomes rather than nominal
pipeline failures.

\CTPrepEndSupplement


\end{document}